%% file: main.tex
\documentclass[lettersize,journal]{IEEEtran}
\usepackage{amsmath,amsfonts}
\usepackage{algorithmic}
\usepackage{algorithm}
\usepackage{array}
\usepackage[caption=false,font=normalsize,labelfont=sf,textfont=sf]{subfig}
\usepackage{textcomp}
\usepackage{stfloats}
\usepackage{url}
\usepackage{verbatim}
\usepackage{graphicx}
\usepackage{cite}
\usepackage{graphicx}
\usepackage{amsmath}
\usepackage{amssymb}
\usepackage{booktabs}
\usepackage{array,multirow,graphicx}
\usepackage{pifont}
\usepackage{setspace}
\usepackage{bbm}
\usepackage{caption}

\usepackage{color,soul}
\newcommand{\XX}[1]{\textcolor{black}{#1}}
\newcommand{\FH}[1]{\textcolor{black}{#1}}

\begin{document}

\title{Self-supervised Domain-agnostic Domain Adaptation for Satellite Images}

\author{Fahong Zhang, Yilei Shi,~\IEEEmembership{Member,~IEEE,} and \XX{Xiao Xiang} Zhu,~\IEEEmembership{Fellow,~IEEE,}
\thanks{
F. Zhang and X. Zhu are with the Chair of Data Science in Earth Observation, Technical University of Munich, 80333 Munich, Germany (e-mail: (fahong.zhang, xiaoxiang.zhu)@tum.de)..
}
\thanks{Y. Shi is with the Chair of Remote Sensing Technology, Technical University of Munich (TUM), 80333 Munich, Germany
(e-mail: yilei.shi@tum.de)}
\thanks{(Correspondence: Xiao Xiang Zhu)}}

\markboth{Journal of \LaTeX\ Class Files,~Vol.~14, No.~8, April~2022}%
{Shell \MakeLowercase{\textit{et al.}}: A Sample Article Using IEEEtran.cls for IEEE Journals}


\maketitle

\begin{abstract}
\input{content/abstract.tex}

\end{abstract}


\section{Introduction}
\input{content/introduction}

\section{Related Works}
\input{content/related_works}

\section{Methods}
\input{content/methods}

\section{Experiments}
\input{content/experiments}

\section{Conclusion}
\input{content/conclusion}

\bibliographystyle{IEEEtran}
\bibliography{mybib}

\end{document}

%% file: content/abstract.tex
\XX{Domain shift caused by, e.g., different geographical regions or acquisition conditions is a common issue in machine learning for global scale satellite image processing. A promising method to address this problem is domain adaptation, where the training and the testing datasets are split into two or multiple domains according to their distributions, and an adaptation method is applied to improve the generalizability of the model on the testing dataset.}
However, defining the domain to which each \XX{satellite} image \XX{belongs} is not trivial,    
especially under large-scale multi-temporal and multi-sensory scenarios,
where a single image mosaic could be generated from multiple data sources.
In this paper, we propose an self-supervised domain-agnostic domain adaptation (SS(DA)$^2$) method to perform domain adaptation without such \XX{a} domain definition.
To achieve this, we first design a contrastive generative adversarial loss to train a generative network to perform image-to-image translation between any two satellite image patches.
Then, we improve the generalizability of the downstream models by augmenting the training data with different testing spectral characteristics.
The experimental results on public benchmarks verify the effectiveness of SS(DA)$^2$.

%% file: content/introduction.tex
\FH{
It is well known that satellite images taken from different locations, with different sensors, or at different times, generally exhibit large spectral variations due to differences in atmospheric conditions, viewing angles, illumination conditions, and so on.
As a result,
a supervised image processing-based model will suffer from a performance decay when it is applied on scenarios where unseen spectral characteristics of the images exist.
A promising approach to tackle this problem and improve the generalizability of the model is to adapt the model trained on existing annotated data to the domains of different data sources.
Such an idea is usually termed as domain adaptation (DA).
}

\begin{figure}[htp]
    \centering
    \includegraphics[width=1.0\linewidth]{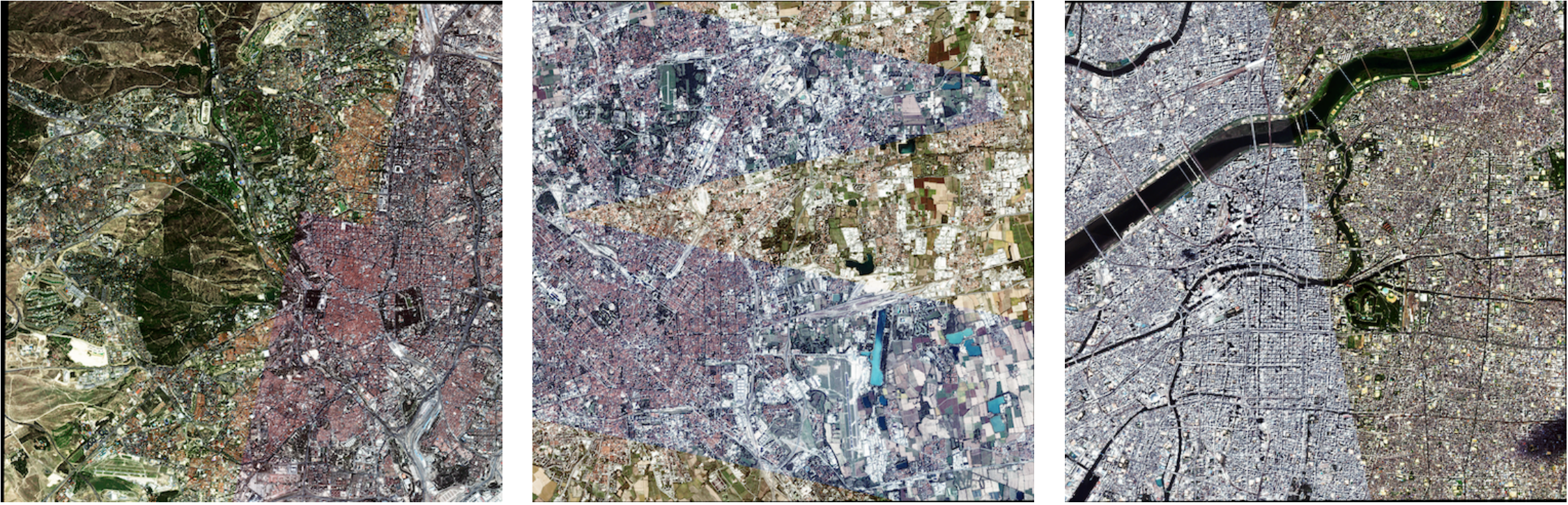}
    \caption{Examples of satellite image mosaics. (From PlanetScope data.)
    }
    \label{fig:mosaics}
\end{figure}

A basic prerequisite for applying conventional DA approach is that the well-defined domain knowledge should be acquired,
i.e.,
knowledge about how to separate the data into multiple domains,
in which a certain level of intra-domain homogeneity and inter-domain heterogeneity \XX{is fulfilled}. 
In specific cases, such domain definition \XX{is naturally given} 
according to the sensory, spatial or temporal information,
especially when the study area is limited.
However, the domain definition becomes non-trivial when we are aiming at \XX{global-scale applications. }
\XX{Fig. \ref{fig:mosaics} exemplifies single image mosaics generated from multi-temporal multi-sensory sources to overcome the limited swath width, cloud coverage, or other limitations.}
As a result, different parts of the mosaic could have different spectral characteristics, which makes it difficult to well define \XX{domains}. 
Although data harmonization techniques have been proposed to mitigate this problem,
their performances on large-scale complicated scenes still remain limited \cite{cresson2015natural}.

\XX{With such considerations, we aim at developing a general DA approach for satellite imagery independent of downstream applications, without relying on any predefined domain separation rule. We term this as domain-agnostic DA.}
Inspired by \cite{tasar2020daugnet},
\XX{we develop} 
a deep learning-based image-to-image translation (I2I) \cite{zhu2017unpaired} method and apply it as \XX{data augmentation to the downstream task-specific models.} 
In this paper, we propose a self-supervised domain-agnostic domain adaptation (SS(DA)$^2$) method to achieve such an I2I without a prior domain definition.
\begin{itemize}
    \item We elaborate and investigate the domain-agnostic domain adaptation problem and propose a general I2I approach to tackle such a problem.
    
    \item We integrate the contrastive learning techniques \cite{chen2020simple} into the adversarial learning pipeline and enable the I2I between two arbitrary image patches
    without explicitly modeling their domain characteristics.
    
    \item The experimental results show that the proposed SS(DA)$^2$ approach can outperform the state-of-the-art I2I-based DA approach without \XX{a domain definition}. 
\end{itemize}

%% file: content/related_works.tex
\begin{figure*}[htp]
    \centering
    \includegraphics[width=0.8\linewidth]{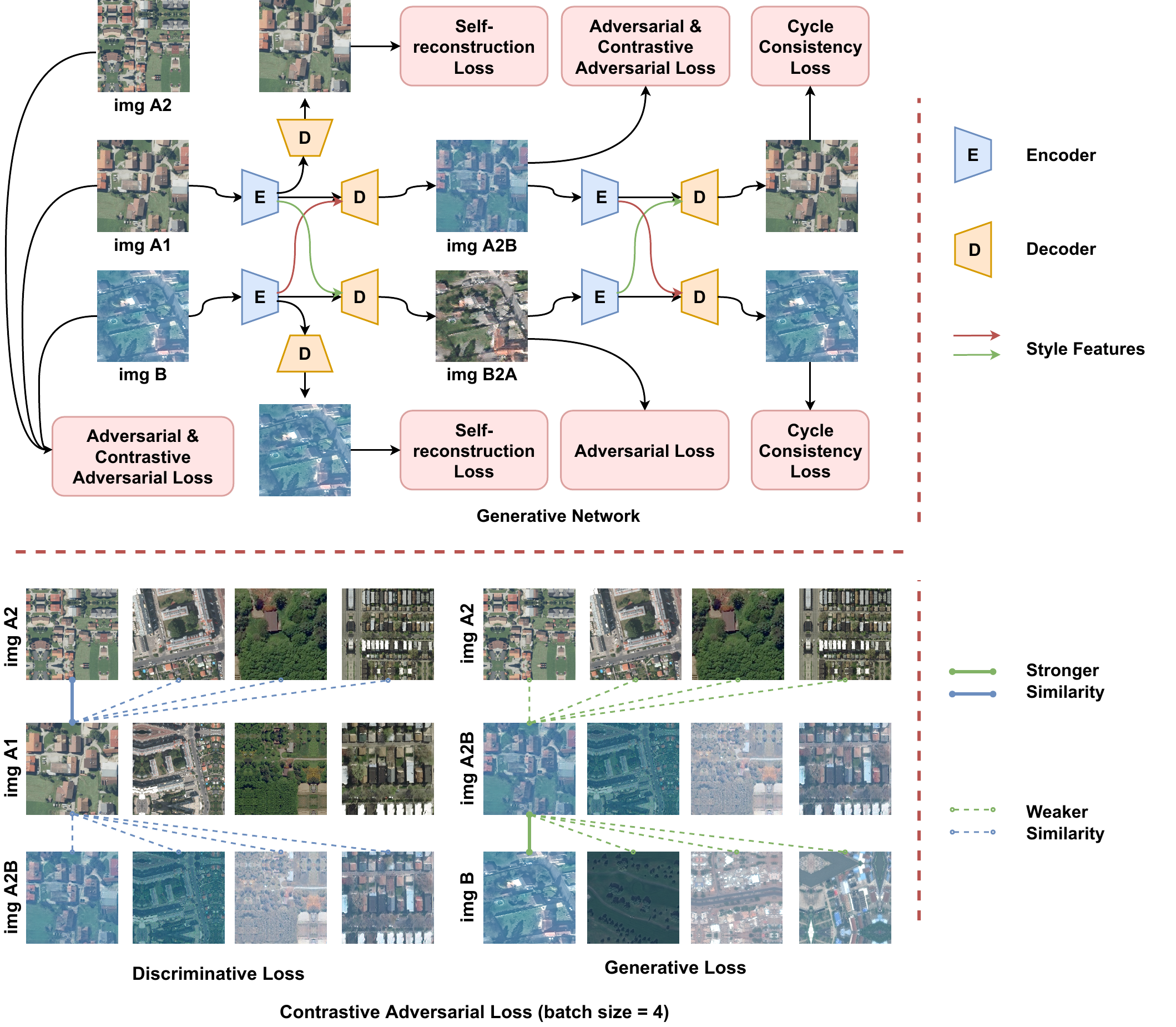}
    \caption{
        Illustration of the overall architecture and the proposed contrastive adversarial loss.
        The inputs to the the network are two randomly sampled images img A and img B, where img A is further augmented twice and derives img A1 and img A2.
        Then, img A1 and img B will be \FH{encoded and merged} to generate the translated images img A2B and img B2A, according to AdaIn \cite{huang2017arbitrary}. Self-reconstruction and cycle consistency loss are applied to ensure the extracted features and the translated images maintain the structural and content information.
        Adversarial loss is utilized to enhance the genuineness of the translated images.
        Contrastive adversarial loss is the key loss function that enables the style transfer.
        More details about it will be discussed in Sec. \ref{sec:contrastive_loss}.
    }
    \label{fig:architecture}
\end{figure*}

\subsection{Image-to-image Translation}
I2I methods aim at learning a mapping function that can map the images sampled from the source domain to the target domain, \FH{ensuring that the mapped images have similar distributions with the target data}.
\FH{Earlier methods are mostly hand-crafted, with the goal to reduce the visual differences between the source and the target images.} Such methods include histogram matching \cite{castleman1996digital}, graph matching \cite{tuia2012graph} and other standardization-based approaches such as histogram equalization \cite{castleman1996digital} and gray world \cite{buchsbaum1980spatial}.

To better exploit the distinctions between different data sources,
data-driven methods and deep learning techniques has been widely explored
to improve the adaptiveness and robustness of the I2I approaches.
Theses approaches are mostly based on the CycleGAN architecture \cite{zhu2017unpaired},
where two generators and two discriminators are trained to map the source data to the target domain and vise versa.
With the development of style transfer techiniques,
Adaptive Instance Transformation (AdaIn) \cite{huang2017arbitrary,huang2018multimodal} is developed to model the style or spectral characteristics of the input images,
saving the need of training separated generative and discriminative models for both the source and target domains.

To tackle more general cases, \FH{especially when} the training and testing data are sampled from multiple domains,
StandardGAN \cite{tasar2020standardgan} designs a network model that can perform I2I \FH{between two arbitrary domains}
by training multiple style encoders for each domain.
In DAug \cite{tasar2020daugnet}, a more simplified model is proposed,
where the style characteristics of different domains are modeled as trainable feature vectors,
which further reduce the computational burdens.
Other multi-domain I2I models include starGAN \cite{choi2018stargan} and starGANv2 \cite{choi2020stargan}.

\subsection{Self-supervised Learning}
Self-supervised learning (SSL) is a branch of unsupervised learning,
where the training data is automatically labeled by exploiting their inward relationships \cite{liu2021self}.
Among all different categories of SSL approaches,
\FH{contrastive learning has demonstrated great potential
by measuring the similarity between sample pairs.
During the measurement learning process,}
Noise Constrastive Estimation (NCE) \cite{gutmann2010noise} or InfoNCE \cite{van2018representation} objectives
will be optimized so that different augmented views of a single sample 
are more similar than views from different samples.

%% file: content/methods.tex
The overall architecture of the proposed SS(DA)$^2$ is illustrated in Fig. \ref{fig:architecture}.

\subsection{Problem Formulation}
Here we first formulate the domain-agnostic I2I problem.
Given a training dataset $\mathcal{D}_{train}$
and a testing dataset $\mathcal{D}_{test}$.
We assume both $\mathcal{D}_{train}$ and $\mathcal{D}_{test}$
consist of multiple domains,
yet the domain assignment knowledge is missing.
Given two randomly sampled images $I_{A}$ and $I_{B}$ from $\mathcal{D}_{train} \cup \mathcal{D}_{test}$,
the goal of I2I is to generate an image $I_{A\rightarrow B}$,
ensuring that its content or spatial geometry is identical to $I_{A}$,
while style or spectral characteristic is similar to $I_{B}$.

SS(DA)$^2$ contains a generator and a discriminator $F$,
where the generator consists of an encoder $E$ and a decoder $D$.
$E$ maps a sampled image to a feature vector: $x = E(I)$,
while $D$ decodes a feature vector to generate an image: $I = D(x)$.
To conduct the self-supervised learning,
\FH{$I_{A}$ will be augmented twice to two different views, including $I_{A_{1}}$ and $I_{A_{2}}$} by random resizing, cropping, and gaussian blurring.
The input to SS(DA)$^2$ during each training step is a batch of randomly sampled $I_{A_{1}}$, $I_{A_{2}}$ and $I_{B}$.

The overall loss functions of SS(DA)$^2$ are given as:
\begin{align}
    \mathcal{L}_{gen} &= \lambda_1 \mathcal{L}_{rec}
    + \lambda_2 \mathcal{L}_{adv}^{gen}
    + \lambda_3 \mathcal{L}_{cyc}
    + \lambda_4 \mathcal{L}_{per}
    + \lambda_5 \mathcal{L}_{con}^{gen},
    \label{Eq:1}
    \\
    \mathcal{L}_{dis} &=
    \lambda_2 \mathcal{L}_{adv}^{dis}
    + \lambda_5 \mathcal{L}_{con}^{dis}.
\end{align}
Here $\mathcal{L}_{gen}$ and $\mathcal{L}_{dis}$ correspond to the losses of the generator and the discriminator, respectively.
They will be optimized alternatively during training.
In the remaining part of this section,
we will introduce each loss item in details.

\subsection{Self-reconstruction Loss}
The self reconstruction loss is applied to ensure that the extracted feature of an image can be used to reconstruct itself.
It can be formulated as:
\begin{equation}
    \mathcal{L}_{rec} = L_{1}(D(x_{A_{1}}), I_{A_{1}}) + L_{1}(D(x_{B}), I_{B}),
\end{equation}
where $L_{1}(\cdot)$ is the smooth $l_{1}$ loss function.

\subsection{Adversarial Loss}
Adversarial loss is applied to make sure the translated images perceptually similar the the genuine ones.
First, the translated images are achieved by decoding the features from two input sources according to AdaIn \cite{huang2017arbitrary}:

\begin{align}
\begin{split}
    x_{A_{1}\rightarrow B} &= AdaIn(x_{A_{1}}, x_{B}), \\
    x_{B\rightarrow A_{1}} &= AdaIn(x_{B}, x_{A_{1}}), \\
    I_{A_{1}\rightarrow B} &= D(x_{A_{1}\rightarrow B}), \\
    I_{B\rightarrow A_{1}} &= D(x_{B\rightarrow A_{1}}).
\end{split}
\end{align}
The adversarial loss is then applied on $I_{A_{1} \rightarrow B}$ and $I_{B \rightarrow A_{1}}$:
\begin{align}
    \begin{split}
    \mathcal{L}_{adv}^{dis}
    &= (F(I_{A_{1}}) - 0)^2 + 
    (F(I_{B}) - 0)^2 \\
    & \ \ \ + (F(I_{A_{1}\rightarrow B}) - 1)^2 + 
    (F(I_{B \rightarrow A_{1}}) - 1)^2.
    \end{split}
\end{align}
\begin{align}
    \begin{split}
        \mathcal{L}_{adv}^{gen}
        = (F(I_{A_{1} \rightarrow B}) - 0)^2
        + (F(I_{B \rightarrow A_{1}}) - 0)^2.
    \end{split}
\end{align}
By minimizing $\mathcal{L}_{adv}^{dis}$,
the discriminator learns to distinguish the generated images from the real ones.
\FH{By minimizing $\mathcal{L}_{adv}^{gen}$,
the generator learns} to cheat the discriminator to believe that the generated images are real.
To this end, optimizing them alternatively will improve the quality of the generated images.

\subsection{Cycle Consistency Loss}
The idea of cycle consistency loss is originally proposed in \cite{zhu2017unpaired}.
It aims to maintain the original structural and content information in the translated images.
First, the generated $I_{A_{1}\rightarrow B}$ and $I_{B \rightarrow A_{1}}$ will be used to reconstruct $I_{A_{1}}$ and $I_{B}$ according to AdaIn:
\begin{align}
    \begin{split}
    x_{A_{1} \rightarrow B \rightarrow A_{1}}
    &= AdaIn(E(I_{A_{1} \rightarrow B}), E(I_{B \rightarrow A_{1}})),\\
    x_{B \rightarrow A_{1} \rightarrow B}
    &= AdaIn(E(I_{B \rightarrow A_{1}}), E(I_{A_{1} \rightarrow B})). \\
    \end{split}
\end{align}
Then, the differences between the reconstructed and the original images will be minimized:
\begin{align}
    \begin{split}
    \mathcal{L}_{cyc}
    &= L_1(D(x_{A_{1} \rightarrow B \rightarrow A_{1}}), I_{A_{1}})\\
    & \ \ \ + L_1(D(x_{B \rightarrow A_{1} \rightarrow B}), I_{B}).
    \end{split}
\end{align}

\subsection{Perceptual Loss}
Perceptual loss \cite{johnson2016perceptual} $\mathcal{L}_{per}$ is used to reduce the high-level perceptual differences between reconstructed and original images.
It can be formulated as:
\begin{align}
\begin{split}
    \mathcal{L}_{per} = \ & \:
    \lvert| E_{per}(D(x_{A_{1}})) - E_{per}(I_{A_{1}})\rvert|_2^2 \\
    &+ \lvert| E_{per}(D(x_{B})) - E_{per}(I_{B})\rvert|_2^2 \\
    &+ \lvert| E_{per}(D(x_{A_{1} \rightarrow B \rightarrow A_{1}})) - E_{per}(I_{A_{1}})\rvert|_2^2 \\
    &+ \lvert| E_{per}(D(x_{B \rightarrow A_{1} \rightarrow B})) - E_{per}(I_{B})\rvert|_2^2,
\end{split}
\end{align}
where $E_{per}$ is a VGG based loss network \cite{simonyan2014very} pretrained on the ImageNet dataset.
    
\subsection{Contrastive Adversarial Loss}
\label{sec:contrastive_loss}
Contrastive adversarial loss is the core
supervision signal that makes the translated images $I_{A_{1} \rightarrow B}$ and $I_{B \rightarrow A_{1}}$ have similar style to $I_{B}$ and $I_{A_{1}}$, respectively.
To implement this, we train the discriminator $F$ as a similarity measurement based on contrastive learning,
and meanwhile optimize the generator in a contrastive and adversarial manner,
so as to cheat $F$ to believe the generated $I_{A_{1} \rightarrow B}$ is similar to $I_{B}$.
Specifically, the contrastive adversarial loss for the generator $\mathcal{L}_{con}^{gen}$ and the discriminator $\mathcal{L}_{con}^{dis}$ can be formulated as:

\begin{align}
\begin{split}
    \mathcal{L}_{con}^{dis} = -\frac{1}{N}\sum_{i=1}^{N} \mathop{log} 
    \frac{e^{S(I_{A_{1}}^{i}, I_{A_{2}}^{i})}}
    {\sum_{j=1}^{N} e^{S(I_{A_{1}}^{j}, I_{A_{2}}^{j})}
    + e^{S(I_{A_{1}}^{j}, I_{A_{1} \rightarrow B}^{j})}},
\end{split}
\end{align}
and
\begin{align}
\begin{split}
    \mathcal{L}_{con}^{gen} = -\frac{1}{N}\sum_{i=1}^{N} \mathop{log} 
    \frac{e^{S(I_{A_{1} \rightarrow B}^{i}, I_{B}^{i})}}
    {\sum_{j=1}^{N} e^{S(I_{A_{1} \rightarrow B}^{j}, I_{B}^{j})}
    + e^{S(I_{A_{1} \rightarrow B}^{j}, I_{A_{2}}^{j})}}.
\end{split}
\end{align}
Here $N$ is the batch size,
and $S(\cdot, \cdot)$ is a similarity metric derived from the discriminator $F$, calculated as the cosine similarity of $F$'s last layer features.

As illustrated in the bottom part of Fig. \ref{fig:architecture},
since $I_{A_{1}}^i$ and its corresponding $I_{A_{2}}^i$ (or img A1 and img A2 in Fig. \ref{fig:architecture}) are derived from the same image with only spatial data augmentation \FH{(i.e., no color distortion applied)},
they should have similar styles or spectral characteristics.
By training the discriminator to measure them as highly similar pair,
the discriminator will \FH{implicitly} learn to measure the domain-level similarity.
Accordingly, by training the generator towards being able to cheat the discriminator to believe that $I_{A_{1} \rightarrow B}^i$ is similar to its corresponding $I_{B}^i$,
the translated $I_{A_{1} \rightarrow B}^i$ will be more and more similar to $I_{B}^i$ in terms of \FH{the spectral characteristic}.
In this way, we realize the I2I between multiple domains without explicitly defining the domain assignments of input patches.

%% file: content/experiments.tex
\subsection{Experimental Settings}
To evaluate the performance of SS(DA)$^2$,
we set building segmentation as the downstream task.
\FH{Two public benchmarks including Inria \cite{maggiori2017can} and DeepGlobe dataset} \cite{demirdeepglobe} are used for training and testing, respectively.
Inria dataset provides aerial images for $10$ cities with a coverage of $810$ $km^2$ and a resolution of $0.3$ $m$.
In our experiments, the data annotated with binary building masks from the \FH{first $5$} cities are used.
During training, the images are downsampled by a factor of $2$ and cropped to $256 \times 256$ patches with a stride of $128$. In total there are $72,000$ training patches.

DeepGlobe dataset provides the satellite data of $4$ cities,
including $24,586$ patches in total with size $650 \times 650$.
The pan-sharpened RGB images with a resolution of $0.31$ $m$ are used in our experiments.
The provided 16-bit images for each city are coverted to 8-bit by cutting out the top $2\%$ brightest pixels in each channel.
Among them, $200$ randomly selected patches of each city are used for testing, while \FH{the others} are used for training the I2I networks.

As shown in Tab. \ref{tab:iou},
we compare the proposed methods with $6$ comparative approaches.
\FH{Among them}, Baseline is a vanilla segmentation model where no I2I is applied.
Hist. Equ. standardizes the training and testing data based on histogram equalization \cite{castleman1996digital}.
HSV \cite{buslaev2020albumentations}, Gamma \cite{buslaev2020albumentations} and RHM \cite{yaras2021randomized} are data augmentation based methods that improve models' generalizability by shifting the spectral characteristics of the training data.
DAug \cite{tasar2020daugnet} is a multi-source domain adaptation approach that can perform I2I between images sampled from $2$ arbitrary domains,
given their domain assignments.

\subsection{Implementation Details}
\FH{For training the I2I network},
we adopt a four-block architecture for both the discriminator and the encoder of the generator.
\FH{Here each} block contains a stack of a 2D convolution, an instance normalization, and a max pooling layer, followed by a ReLU activation function.
The numbers of channels of these four blocks are 256, 128, 64 and 32.
We adopt a Unet \cite{DBLP:conf/miccai/RonnebergerFB15} architecture for the decoder of the generator,
The batch size is set to $8$, and the learning rate is set to $0.01$ initially, following a polynomial learning rate decay with a power of $0.95$.
The training process lasts for $100,000$ iterations.
For fair comparison, we use the same network architecture and the same training setting when re-implementing the DAug method \cite{tasar2020daugnet}.
The loss weight in Eq. \ref{Eq:1} is set to $\lambda_1 = 50, \lambda_2=5, \lambda_3=50, \lambda_4=1$ and $\lambda_5=1$.

For the downstream semantic segmentation model,
we use a Unet architecture with a ResNet50 \cite{7780459} backbone.
The batch size is set to $8$.
The learning rate is set to $0.001$ initially,
followed by a polynomial learning rate decay with a power of $0.9$.
\FH{For comparative approaches including RHM, DAug and SS(DA)$^2$, as they all require a reference patch or domain id to perform the I2I},
we randomly sample a reference patch from the testing cities for each of them, and perform the I2I with a probability of $0.5$ during \FH{the training phase}.

\begin{figure}[htp]
    \centering
    \includegraphics[width=1.0\linewidth]{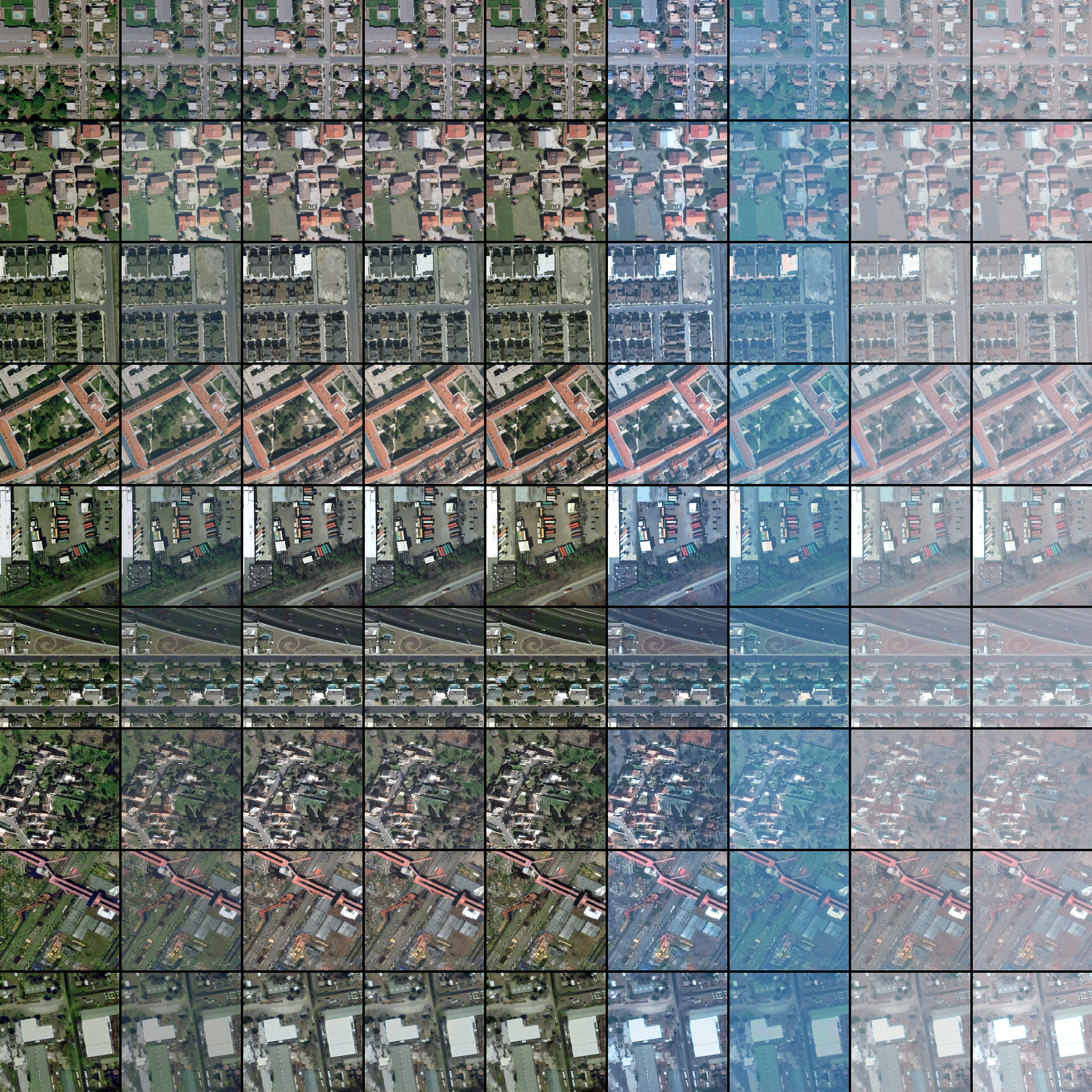}
    \caption{
        Visualization of the images generated by SS(DA)$^2$.
        Diagonal lines of the image matrix are the original images
        sampled from each city.
        The first five images are from Inria dataset,
        while the last four are from DeepGlobe dataset.
        The image located in row $i$ and column $j$ ($i \neq$ j) is generated from the original image in row $i$,
        with the style of the original image in row $j$.
    }
    \label{fig:visualization}
\end{figure}

\setlength\tabcolsep{1pt}
\begin{table}
\footnotesize
\renewcommand\arraystretch{1.2}
\newcolumntype{C}{>{\centering \arraybackslash}m{1.2cm}}
\newcommand\RotText[1]{\rotatebox[origin=c]{60}{\parbox{1.3cm}{\centering#1}}}
\begin{center}
    \begin{tabular}{>{\centering \arraybackslash}m{2.0cm} C C C C | >{\centering \arraybackslash}m{1.3cm}}
\toprule
\addlinespace[1ex]
        Method & Vegas & Paris & Shang. & Khart. & Mean \\
\addlinespace[1ex]
\hline
\addlinespace[1ex]
        Baseline & 60.87 & 43.86 & 45.05 & 43.38 & 48.29 \\
        Hist. Equ. & 62.02 & 49.83 & 45.62 & 19.84 & 44.33 \\ 
        HSV & 69.52 & 49.34 & 47.49 & 38.75 & 51.28\\
        Gamma & 63.35 & 45.50 & 48.56 & \bf{51.64} & 52.26\\
        RHM \cite{yaras2021randomized} & 72.86 & 60.00 & 55.56 & 33.08 & 55.38 \\
        DAug \cite{tasar2020daugnet} & \bf{74.30} & \bf{60.14} & \bf{58.45} & 42.38 & 58.82\\
        SS(DA)$^2$ &  72.59 & 57.33 & 57.49 & 51.36 & \bf{59.69} \\
\addlinespace[0.5ex]
\bottomrule
\end{tabular}
\end{center}
\caption{
    IoU (\%) for the building class on Inria $\rightarrow$ DeepGlobe benchmark.
    Averaged results and \FH{the results for each testing city}, including Vegas, Paris, Shanghai and Khartoum are reported.
}
\label{tab:iou}
\end{table}

\subsection{Quantitative Results}
The quantitative comparative results are listed in Tab. \ref{tab:iou}.
As can be observed,
all the model-based data augmentation approaches can improve over the baseline model in terms of the averaged IoU results.
And it turns out that \FH{a certain augmentation-based} approach could be able to tackle certain \FH{types of the spectral shift}, e.g., Gamma performs quite well on Khartoum, and HSV has a large improvement on Vegas \FH{against the baseline}.
In contrast, histogram equalization based approach produces poor results,
which indicates the large shifts between training and testing data are difficult to be handled by simple standardization \FH{approaches}.
On the other hand, deep learning based I2I methods including DAug and SS(DA)$^2$ generally have more stable improvements against the baseline regarding their performances on different cities,
which proves the superiority of learning-based model in dealing with the complex multi-source domain adaptation setting.
The proposed SS(DA)$^2$ can outperform DAug in terms of the results on Khartoum, and the averaged results over the four cities, despite the lack of domain assignment.
This indicates the robustness of the proposed method,
and further demonstrates that the intra-domain discrepancy can be well tackled by patch-level self-supervised I2I.

\subsection{Qualitative Results}
The qualitative results of SS(DA)$^2$ are listed in Fig. \ref{fig:visualization}.
As can be seen,
images in the first five columns are with relatively consistent styles,
since they are aerial images and do not suffer from atmospheric distortions.
In contrast, the last four columns from DeepGlobe dataset share quite different spectral characteristics to the others,
\FH{which demonstrates the existence of domain shifts and 
highlights the difficulty to perform adaptation among them.}
According to the visualization results,
the translated images in the same column generally have similar styles,
while the images in the same row share similar spatial contents.
This perceptually \FH{verifies} that the proposed SS(DA)$^2$ can successfully capture the spectra-related differences in two input patches and transfer the style between them through self-supervised and adversarial learning.

%% file: content/conclusion.tex
In this paper,
we elaborate the importance of domain-agnostic DA in \XX{machine learning} for large-scale real-world applications,
and propose \XX{a self-supervised approach based} on contrastive learning and adversarial learning techniques,
which realizes the I2I between any pair of randomly sampled patches without pre-defined \XX{domain assignments}.
Experimental results show that the proposed method can outperform the other model-based or learning-based competitors and is able to generate perceptually high-quality images.